\documentclass{article}
\usepackage{spconf,amsmath,graphicx}
\usepackage{textcomp}
\usepackage{boldline}
\usepackage{algorithm}
\usepackage{algorithmic}
\usepackage{booktabs} 
\usepackage{multirow}
\usepackage{multicol} 
\usepackage{changepage}
\usepackage{subfigure}
\usepackage{float}
\usepackage{amssymb}
\usepackage{setspace}
\usepackage{cite}

\title{MPS-AMS: Masked Patches Selection and Adaptive Masking Strategy Based Self-Supervised Medical Image Segmentation}
\name{\begin{tabular}{c}Xiangtao Wang$^{1}$, Ruizhi Wang$^{1}$, Biao Tian$^{1}$, Jiaojiao Zhang$^{1}$, Shuo Zhang$^{1}$, \\ Junyang Chen$^{2}$, Thomas Lukasiewicz$^{3,4}$, Zhenghua Xu$^{1,\dag}$\thanks{$^{\dag}$Corresponding author: zhenghua.xu@hebut.edu.cn (Zhenghua Xu).}\end{tabular}}
\address{
$^1$State Key Laboratory of Reliability and Intelligence of Electrical Equipment, School of Health\\ Sciences and Biomedical Engineering, Hebei University of Technology, Tianjin, China\\
$^2$College of Computer Science and Software Engineering and Guangdong Laboratory
of Artificial\\ Intelligence and Digital Economy (SZ), Shenzhen University, Shenzhen, China\\
$^3$Institute of Logic and Computation, TU Wien, Vienna, Austria\\
$^4$Department of Computer Science, University of Oxford, Oxford,   United Kingdom\\
}
\begin{document}
%
\maketitle
\begin{abstract}
Existing self-supervised learning methods based on contrastive learning and masked image modeling have demonstrated impressive performances. However, current masked image modeling methods are mainly utilized in natural images, and their applications in medical images are relatively lacking. Besides, their fixed high masking strategy limits the upper bound of conditional mutual information, and the gradient noise is considerable, making less the learned representation information. Motivated by these limitations, in this paper, we propose masked patches selection and adaptive masking strategy based self-supervised medical image segmentation method, named MPS-AMS. We leverage the masked patches selection strategy to choose masked patches with lesions to obtain more lesion representation information, and the adaptive masking strategy is utilized to help learn more mutual information and improve performance further. Extensive experiments on three public medical image segmentation datasets (BUSI, Hecktor, and Brats2018) show that our proposed method greatly outperforms the state-of-the-art self-supervised baselines.
\end{abstract}
\vspace{-0.5em}
\begin{figure*}
    \centering
    \vspace{-2em}
\includegraphics[width=0.9\textwidth]{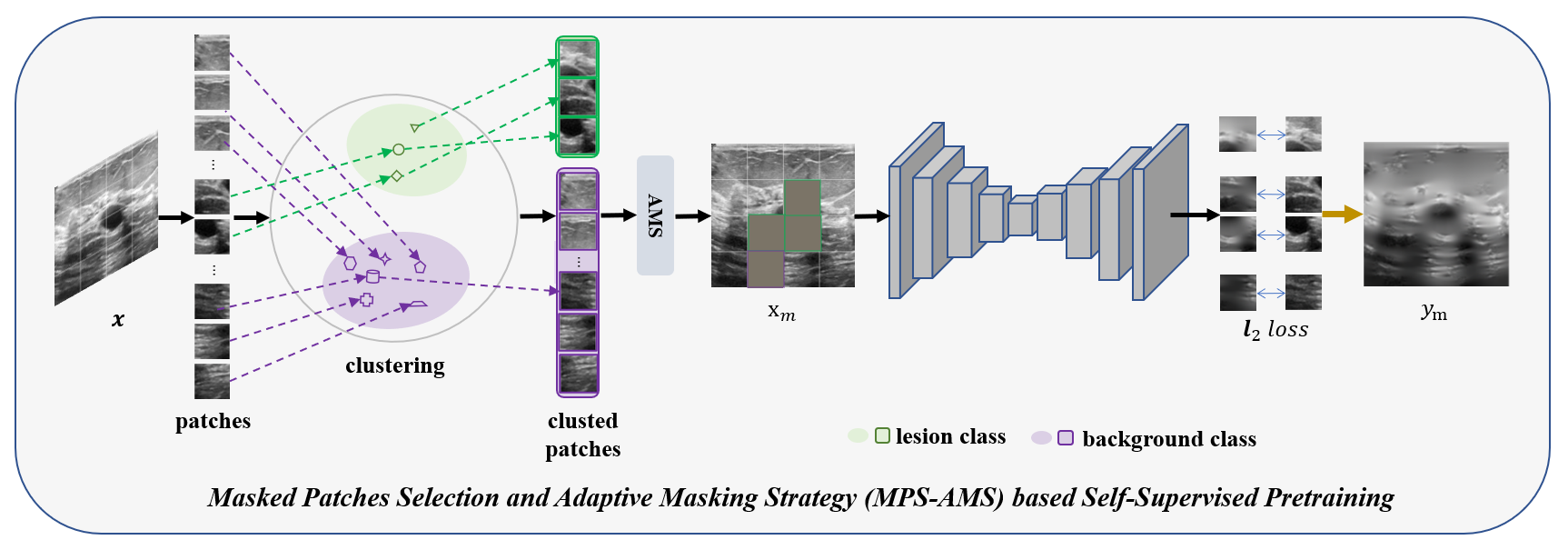}
        \vspace{-1.5em}
        \caption{The illustration of our MPS-AMS architecture. Green and purple areas represent lesion and background, respectively.
        }
        \vspace{-1.5em}
        \label{fig:network} 
\end{figure*}

\vspace{-1em}
\section{Introduction}
\label{sec:intro}
\vspace{-0.5em}

Deep learning has demonstrated remarkable achievements in medical image analysis~\cite{w-Net2022,DRLHomoPre2022}.
In particular, self-supervised learning (SSL) has emerged as a crucial technique for medical image segmentation tasks~\cite{chen2019self,adaptive_2}, which is mostly based on contrastive learning. Contrastive learning~\cite{simCLR,BYOL,SwAV} enforces positive samples closer and negative samples further away in latent space to learn representation information. However, these methods only focus on the global semantics of the image and ignore the details of the image and non-subject areas~\cite{CAE}.
To solve these problems, masked image modeling~\cite{MAE, Simmim, MixMIM, Uniform_Masking} for self-supervised pretraining has come into being and recently grown in popularity. Masked image modeling (MIM) aims to reconstruct corresponding discrete visual tokens from masked input, like MAE~\cite{MAE} and SimMIM~\cite{Simmim}. MAE leverages an asymmetric encoder and decoder architecture to predict masked patches from unmasked ones directly. To further maintain image structure, SimMIM takes visible and masked patches as input, and it also lightweights decoder to accelerate pretraining process.

Although MAE and its variants~\cite{Simmim, MixMIM, Uniform_Masking}  have shown promising results, their strategies for selecting masked patches and masking ratio are still unsatisfactory. Specifically, they have not been extensively applied in medical images, where the lesion area is usually small and may be overlooked, resulting in less lesion representation information and limiting the performance of downstream tasks. Additionally, a fixed high masking rate leads to a small learnable conditional mutual information and large gradient noise, which lowers the upper bound of representation information learned and makes optimization challenging\cite{mutual_information,adaptive_1}. Therefore, the need for masked patches selection and adaptive masking strategy in medical images is compelling.

In this paper, we innovatively propose \textbf{M}asked \textbf{P}atches \textbf{S}election and \textbf{A}daptive \textbf{M}asking \textbf{S}trategy based self-supervised medical image segmentation \textbf{(MPS-AMS)}. First, we leverage the masked patches selection strategy to focus on lesions to learn more lesion representation information, which is achieved by choosing the masked patches with a high probability of containing lesions through covariance matrix and k-means clustering. Then, we propose an adaptive masking ratio strategy to improve the upper bound of conditional mutual information to learn more representation information.

The contributions of this paper are briefly summarized as follows: (i) We propose a novel masked patches selection strategy specifically for medical images and an adaptive masking strategy to overcome the shortcomings of existing masked image modeling methods. (ii) To enhance the lesion representation information, we use the masked patches selection strategy to select patches with a higher probability of containing lesions and the adaptive masking ratio strategy to reduce gradient noise and improve the upper bound of conditional mutual information. (iii) Extensive experiments on three public medical image datasets demonstrate that MPS-AMS outperforms state-of-the-art self-supervised methods, and the proposed strategies are effective and essential for improving model performance.

\vspace{-1em}
\section{METHODOLOGY}
\label{sec:method}
\vspace{-0.5em}

Figure~\ref{fig:network} illustrates the overall structure of our proposed MPS-AMS, which comprises two main processing steps. Firstly, MPS-AMS conducts masked image modeling pretraining using a large set of unlabeled medical images. The resulting modules are then 
utilized in fully supervised downstream segmentation tasks with a small amount of labeled images. 
%
\begin{small}
    \begin{table*}[!t]
        	\vspace{-2em}
        \small
        \begin{center}
            \caption{Results of the proposed MPS-AMS and baselines on BUSI, Hecktor, and Brats2018 datasets.}
            \label{table:MainResults}
            \vspace{0.35em}
            \begin{tabular}{m{0.06\textwidth}<{\centering}|m{0.06\textwidth}<{\centering}|m{0.06\textwidth}<{\centering}|m{0.06\textwidth}<{\centering}|m{0.06\textwidth}<{\centering}|m{0.06\textwidth}
            <{\centering}|m{0.06\textwidth}<{\centering}|m{0.06\textwidth}<{\centering}|m{0.06\textwidth}<{\centering}|m{0.06\textwidth}<{\centering}|m{0.06\textwidth}<{\centering}}
                \hlineB{3}
                \multicolumn{2}{c|}{\multirow{2}{*}{\textbf{Methods}}} 
                & \multicolumn{3}{c|}{\multirow{1}{*}{\textbf{BUSI}}}
                & \multicolumn{3}{c|}{\multirow{1}{*}{\textbf{Hecktor}}} 
                & \multicolumn{3}{c}{\multirow{1}{*}{\textbf{Brats2018}}} \\  
                \cline{3-11}
            
                \multicolumn{2}{c|}{}  
                &{DSC}&{PPV}&{Sen}
                &{DSC}&{PPV}&{Sen} 
                &{DSC}&{PPV}&{Sen} \\ \hlineB{3}
                
                \multirow{8}{*}{\textbf{5\%}}{}&\multicolumn{1}{c|}{U-Net} 
                & {0.3863} & {0.5234} & {0.4531}
                & {0.1762} & {0.2803} & {0.1755}
                & {0.2059} & {0.2253} & {0.2606}
                \\ 
                
                \multicolumn{1}{c|}{}&\multicolumn{1}{c|}{SimCLR} 
                & {0.4172} & {0.4129} & {0.3554} 
                & {0.2201} & {0.2385} & {0.3113} 
                & {0.2908} & {0.3009} & {0.4376}\\ 

                \multicolumn{1}{c|}{}&\multicolumn{1}{c|}{BYOL} 
                & {0.4291} & {0.6991} & {0.4311} 
                & {0.1967} & {0.2179} & {0.2555} 
                & {0.2811} & {0.2867} & {0.4545}\\ 
                
                \multicolumn{1}{c|}{}&\multicolumn{1}{c|}{SwAV} 
                & {0.4017} & {0.6128} & {0.4470}
                & {0.2186} & {0.1909} & {\textbf{0.3793}} 
                & {0.2277} & {0.1884} & {0.4466}\\ 

                \multicolumn{1}{c|}{}&\multicolumn{1}{c|}{MAE} 
                & {0.4793} & {0.6568} & {0.5463}
                & {0.2560} & {\textbf{0.2975}} & {0.2966} 
                & {0.2898} & {0.3012} & {0.4596}\\ 
                
                \multicolumn{1}{c|}{}&\multicolumn{1}{c|}{SimMIM} 
                & {0.4644} & {0.6847} & {0.4951}
                & {0.2413} & {0.2745} & {0.2972}
                & {0.2801} & {\textbf{0.3095}} & {0.4297}\\ 
                
                \multicolumn{1}{c|}{}&\multicolumn{1}{c|}{\textbf{MPS-AMS}} 
                & {\textbf{0.5002}} & {\textbf{0.7034}} & {\textbf{0.5661}} 
                & {\textbf{0.2711}} & {\textbf{0.2975}} & {0.3347}
                & {\textbf{0.2973}} & {0.3035} & {\textbf{0.4708}}  \\ \hlineB{3}

                \multirow{8}{*}{\textbf{10\%}}{}&\multicolumn{1}{c|}{U-Net} 
                & {0.4876} & {0.6360} & {0.5262}
                & {0.2541} & {0.3002} & {0.2875}
                & {0.2529} & {0.2677} & {0.3366}
                \\ 
                
                \multicolumn{1}{c|}{}&\multicolumn{1}{c|}{SimCLR} 
                & {0.5396} & {0.6439} & {0.5759} 
                & {0.2947} & {0.3325} & {0.3900} 
                & {0.3551} & {\textbf{0.3459}} & {0.4868}\\ 

                \multicolumn{1}{c|}{}&\multicolumn{1}{c|}{BYOL} 
                & {0.5491}& {0.7044} & {0.5761} 
                & {0.3013} & {0.3106}  & {0.3930} 
                & {0.3458} & {0.3058} & {0.3535}\\ 
                
                \multicolumn{1}{c|}{}&\multicolumn{1}{c|}{SwAV} 
                & {0.5163} & {0.6325} & {0.5372} 
                & {0.2550} & {0.2669} & {0.3340} 
                & {0.2914} & {0.2562} & {0.4694}\\ 

                \multicolumn{1}{c|}{}&\multicolumn{1}{c|}{MAE} 
                & {0.5639} & {0.6603} & {0.6104}
                & {0.3195} & {0.3443} & {0.3794}
                & {0.3578} & {0.3220} & {0.4878}\\ 

                \multicolumn{1}{c|}{}&\multicolumn{1}{c|}{SimMIM} 
                & {0.5537} & {0.6918} & {\textbf{0.6262}} 
                & {0.2920} & {0.3325} & {0.3511} 
                & {0.3246} & {0.3149} & {0.4725}\\ 
                
                \multicolumn{1}{c|}{}&\multicolumn{1}{c|}{\textbf{MPS-AMS}} 
                & \textbf{{0.5914}} & {\textbf{0.7305}} & {0.6211} 
                & {\textbf{0.3554}} & {\textbf{0.3681}} & {\textbf{0.4125}} 
                & {\textbf{0.3633}} & {0.3163} & {\textbf{0.5019}}\\ \hlineB{3}

                \multirow{1}{*}{\textbf{50\%}}{}&\multicolumn{1}{c|}{U-Net} 
                & {0.5714} & {0.6339} & {0.6058}
                & {0.3090} & {0.3801} & {0.3160} 
                & {0.3535} & {0.3530} & {0.4139} \\ \cline{1-11}  
                \multirow{1}{*}{\textbf{100\%}}{}&\multicolumn{1}{c|}{U-Net} 
                & {0.6821} & {0.8005} & {0.6542}
                & {0.3927} & {0.4523} & {0.4736}  
                & {0.4294} & {0.4497} & {0.5224} \\ \hlineB{3}  
                
            \end{tabular}
        \end{center}
        \vspace{-1.5em}
    \end{table*}
\end{small}

\vspace{-1em}
\subsection{Masked Patches Selection Strategy}
\label{sec:patch_selection}
\vspace{-0.5em}

Since current masked image modeling works are mostly focused on natural images, we first propose masked patches selection strategy special for medical images. We sort all the patches in the order of lesion first, then background, and then mask patches in this order.
The input image $x$ contains two parts, $x=\lbrace x_{i},x_{\neg{i}}\rbrace$, where $x_{i}$ indicates visible patches and $x_{\neg{i}}$ indicates masked patches.

To get the selected $x_{i}$ and $x_{\neg{i}}$, we define two initialized cluster centers, which are predicted to represent lesion and background class respectively. After dividing $x$ into patches, we construct a covariance matrix after a softmax layer according to the degree of similarity between different patches. Then, we take k-means to divide all patches into two categories. Considering that most lesions in medical images only occupy a small overall area, we can suppose the category with a small number of clusters as lesions.  Besides, we choose k-means because it can achieve good performance with lower complexity and faster efficiency compared with other clustering methods like hierarchical, t-SNE, and so on, and the reason why
we choose k-means is well discussed in the section of results.

We evaluate the effectiveness of the proposed masked patches selection strategy by estimating the conditional entropy to represent the uncertainty of the sampling strategy. After classifying patches, the uncertainty of 
$x_{i}$ is reduced, which indicates an improvement of the lower bound.
Concretely, We leverage $H_j$ to indicate the uncertainty of the sampling strategy. $H_1$ is the lowest bound of uncertainty,

\vspace{-0.5em}
\begin{small}
\begin{equation}
    H_1=\mathbb{E}_{p(x_{i},x_{\neg{i}})}logP(x_{i},x_{\neg{i}}).
\end{equation}
\end{small}
\vspace{-1em}

\noindent $H_2$ is used to indicate the uncertainty in the learning process of neural networks,

\vspace{-0.5em}
\begin{small}
\begin{equation}
    H_2=\mathbb{E}_{p}(x_{i},x_{\neg{i}})logQ(x_{i},x_{\neg{i}}).
\end{equation}
\end{small}
\vspace{-1em}

\noindent Suppose $H_3$ is the optimal upper bound with the Monte Carlo sampling strategy,

\vspace{-0.5em}
\begin{small}
\begin{equation}
    H_3=\mathbb{E}_{p}(\hat{x_{i}},\hat{x_{\neg{i}}})logP(x_{i},x_{\neg{i}}).
\end{equation}
\end{small}
\vspace{-1em}

\noindent where $\hat{x_{i}}$ and $\hat{x_{\neg{i}}}$ represent the best sampling results.

The Monte Carlo sampling strategy is shown below, in the interval $[a,b]$, $f(x)$ represents the size of the value, $p(x)$ represents the probability of occurrence, the meaning of the integral result is the output total value.

\vspace{-0.5em}
\begin{small}
\begin{equation}
\int_{a}^{b} h(x) d x=\int_{a}^{b} f(x) p(x) d x=E_{p(x)}[f(x)]
\end{equation}
\end{small}
\vspace{-0.5em}

Because $KL(P \Vert Q) > 0$, we can get $H_2 \leq H_1$. Combining the definitions of conditional entropy described in detail 
~\cite{conditional_entropy}, we can get $H_3 \leq H_2 \leq H_1,$
\noindent which means that our proposed masked patches selection strategy reduces uncertainty and can help to learn more lesion representation information. 

\vspace{-1.0em}
\subsection{Adaptive Masking Strategy}
\label{sec:adaptive_masking}
\vspace{-0.5em}

In contrast to the fixed high masking ratio used in MAE and SimMIM, we propose a novel adaptive masking ratio strategy that combines following insights. Fine-tuning performance is limited under high and fixed masking ratios using in MAE. Furthermore, we note that model's ability to learn representation information and conditional mutual information is higher under larger masking ratios.

Concretely, the initial adaptive ratio $\sigma_0$ is 25\%, which is set according to MAE and it increases with the training process. 

\vspace{-1em}
\begin{small}
\begin{equation}
    \sigma = \sigma_0 + \ln(x_e) / \tau,
\end{equation}
\end{small}
\vspace{-1em}

\noindent where $x_e$ donates the training epoch and $\tau$ is a constant.

\begin{small}
    \begin{table*}[!t]
        \vspace{-1em}
        \small
        \begin{center}
            \caption{Results of our ablation studies on three datasets.}
            \label{table:ablation_study}
            \vspace{0.5em}
            \resizebox{\linewidth}{!}{
            \begin{tabular}{m{0.06\textwidth}<{\centering}|m{0.06\textwidth}<{\centering}|m{0.06\textwidth}<{\centering}|m{0.06\textwidth}<{\centering}|m{0.06\textwidth}<{\centering}|m{0.06\textwidth}
            <{\centering}|m{0.06\textwidth}<{\centering}|m{0.06\textwidth}<{\centering}|m{0.06\textwidth}<{\centering}|m{0.06\textwidth}<{\centering}|m{0.06\textwidth}<{\centering}}
                \hlineB{3}
                \multicolumn{2}{c|}{\multirow{2}{*}{\textbf{Methods}}} 
                & \multicolumn{3}{c|}{\multirow{1}{*}{\textbf{BUSI}}}
                & \multicolumn{3}{c|}{\multirow{1}{*}{\textbf{Hecktor}}} 
                & \multicolumn{3}{c}{\multirow{1}{*}{\textbf{Brats2018}}} \\  
                \cline{3-11}
            
                \multicolumn{2}{c|}{}  
                &{DSC}&{PPV}&{Sen}
                &{DSC}&{PPV}&{Sen} 
                &{DSC}&{PPV}&{Sen} \\ \hlineB{3}

                \multirow{4}{*}{\textbf{5\%}}{}&\multicolumn{1}{c|}{\emph{base}} 
                & {0.4584} & {0.6773} & {0.4841}
                & {0.2370} & {0.2636} & {0.3051}
                & {0.2757} & {0.2302} & {0.4227} \\ 

                \multicolumn{1}{c|}{}&\multicolumn{1}{c|}{\emph{base+AMS}} 
                & {0.4629} & {0.6690} & {0.4498} 
                & {0.2479} & {0.2778} & {0.3087}
                & {0.2790} & {0.2806} & {0.3653}\\ 

                \multicolumn{1}{c|}{}&\multicolumn{1}{c|}{\emph{base+MPS}} 
                & {0.4732} & {\textbf{0.7459}} & {0.4859} 
                & {0.2521} & {0.2865} & { 0.2940}
                & {0.2801} & {\textbf{0.3095}} & {0.4297}\\ 
                
                \multicolumn{1}{c|}{}&\multicolumn{1}{c|}{\emph{\textbf{base+AMS+MPS}}} 
                & {\textbf{0.5002}} & {0.7034} & {\textbf{0.5661}} 
                & {\textbf{0.2711}} & {\textbf{0.2975}} & {\textbf{0.3347}}
                & {\textbf{0.2973}} & {0.3035} & {\textbf{0.4708}}  \\ \hlineB{3}
 
            \end{tabular}}
        \end{center}
        \vspace{-2em}
    \end{table*}
\end{small}

Combining with masked patches selection strategy, we can get the numbers of masked patches $n$ and $x_m$, where $n = \lfloor N \times \sigma \rfloor$ and $x_m$ indicates the image with masked patches. 
As illustrated in Fig~\ref{fig:network}, it is achieved by the $l_2$ loss. 

\vspace{-0.5em}
\begin{small}
\begin{equation}
    L = \sum_{i=1}^{n}({y_{m_i}}-{x_{m_i}})^2,
\end{equation}
\end{small}
\vspace{-0.5em}

\noindent where $x_{m_i}$ indicates the $i$-th masked patch, $y_{m_i}$ indicates the $i$-th reconstructed patch.

\begin{figure*}
    \centering
    \vspace{-3em}
    \includegraphics[width=0.9\textwidth]{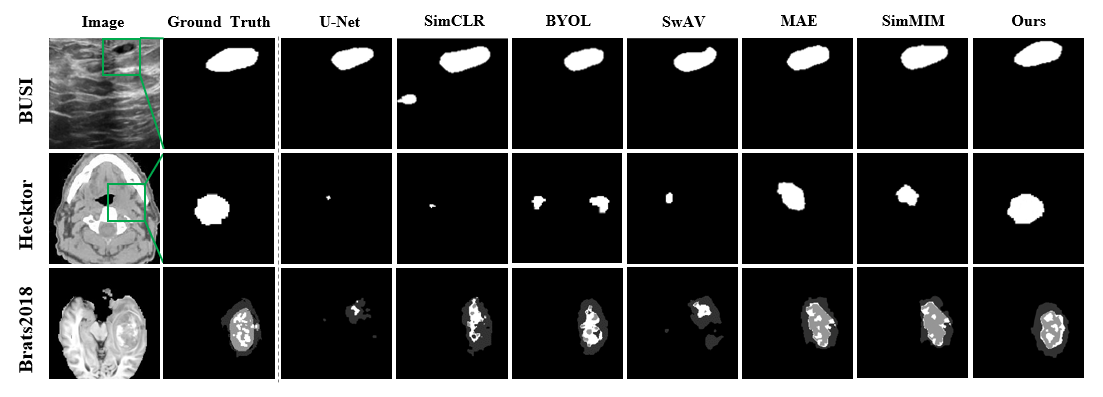}
    \vspace{-1.5em}
    \caption{Visualized segmentation results on the BUSI, Hecktor and BraTS2018 datasets with 10\% labeled data. 
    }
    \label{fig:visualization}
    \vspace{-1em}
\end{figure*}

\vspace{-1em}
\section{EXPERIMENTS AND RESULTS}
\label{sec:result}

\vspace{-0.5em}
\subsection{Experimental settings}
\vspace{-0.5em}

To evaluate the effectiveness of our proposed MPS-AMS, we perform extensive experiments on three publicly medical image datasets with supervised learning and state-of-the-art SSL approaches. The results are shown in Table~\ref{table:MainResults}.

\vspace{-1.5em}
\subsubsection{Datasets}
\vspace{-0.5em}

(i) The BUSI dataset~\cite{BUSI} contains ultrasound scans of breast cancer and consists of $780$ images categorized into normal, benign, and malignant. The average image size is $500\times500$ pixels. (ii) The Hecktor dataset~\cite{Hecktor1, Hecktor2} contains $25923$ slides with CT and PET modalities for head and neck tumor segmentation. (iii) The BraTS2018 dataset~\cite{Brats2018_1,Brats2018_2,Brats2018_3} was released for segmenting brain tumors and includes $22963$ scans with four MRI modalities: T1, T1CE, T2, and FLAIR volumes. Furthermore, we selected CT and T1 modalities for experiments on the Hecktor and Brats2018 datasets as they are challenging to segment and can demonstrate the effectiveness of the proposed method on complex datasets.

\vspace{-1.5em}
\subsubsection{Implementation details}
\vspace{-0.5em}

Our MPS-AMS is implemented based on Torch 1.7.0 and CUDA-10.1. For pretraining, we employ ADAM~\cite{adam} optimizer with a learning rate of $0.0002$, and the batch size is $32$ for BUSI, $48$ for Hecktor, and $36$ for BraTS2018. We set $\tau$ to $12$ to ensure a final masking ratio of nearly $80\%$. For transfer learning, we use the U-Net~\cite{UNet} for downstream segmentation task with ADAM optimizer, an initial learning rate of $0.0002$, weight decay of $0.0001$, and the learning rate strategy is warmup-cosine-lr. The batch size is set to $32$, $31$, and $56$ for datasets with $5\%$ labeled data, and $32$, $90$, and $70$ for datasets with $10\%$ labeled data, respectively. All datasets are randomly divided by $8$:$1$:$1$. The training epochs are set to $200$ for contrastive learning methods, $800$ for masked image modeling methods, and $70$ for fine-tuning. The experiments are conducted on $8$ GeForce RTX2080 GPUs.

\vspace{-1.5em}
\subsubsection{Evaluation}
\vspace{-0.5em}

We employ three widely used metrics to evaluate our method, including positive predictive value (PPV), sensitivity (Sen), and dice similarity coefficient (DSC). PPV is defined as the ratio of correctly segmented positive pixels to all pixels classified as positive in the segmentation result. Sen represents the ratio of correctly segmented positive pixels to all pixels annotated as positive in the ground truth. DSC is the harmonic mean of PPV and Sen, providing a more comprehensive assessment of model performance.

\vspace{-1.5em}
\subsubsection{Baselines}
\vspace{-0.5em}

To evaluate the performance of our proposed MPS-AMS, we choose randomly initialized U-Net without self-supervised pretraining as the full-supervised baselines, leveraging $5\%$ and $10\%$ annotations ratio. 
Besides, several state-of-the-art self-supervised learning methods are chosen as the self-supervised learning baselines in our experiments, including SimCLR~\cite{simCLR}, BYOL~\cite{BYOL}, SwAV~\cite{SwAV}, MAE~\cite{MAE}, and SimMIM~\cite{Simmim}. We evaluate the quality of the learned representations by transferring the weight from different self-supervised learning methods to the medical image segmentation task, and then we evaluate their downstream performances.

All baselines are implemented and run leveraging similar procedures and settings as those in their original papers, and additional parameter adjustments are made to our best efforts.

\vspace{-1em}
\subsection{Main results}
\vspace{-0.5em}

To investigate the effectiveness of MPS-AMS, we conduct experiments on three datasets and compare the performance with two state-of-the-art baselines: Fully Supervised Baseline (i.e., Fully Supervised) and Self-Supervised Baselines (i.e., SimCLR, MAE). For a fair comparison, we use the same backbone network (U-Net) with $5\%$ and $10\%$ annotations across all methods. The experimental results are shown in Table~\ref{table:MainResults} and the segmentation results are shown in Figure~\ref{fig:visualization}.

As shown in Table~\ref{table:MainResults} and Table~\ref{table:ablation_study}, MPS-AMS generally outperforms all baselines, which proves it achieves better segmentation performance with limited annotations. Besides, our proposed strategies are also well demonstrated.

\textbf{Compare with Fully Supervised Learning from Scratch}.
Specifically, MPS-AMS generally outperforms the baseline model trained from scratch by a large margin with $5\%$ and $10\%$ annotations. Furthermore, when using $10\%$ annotations, we can generally outperform the fully supervised method with $50\%$ annotations.


\begin{small}
    \begin{table}[!t] 
        \caption{Results of different clustering methods on Brats2018 with 10\% labeled data.}
        \vspace{0.3em}
        \label{table_clustering}
        \begin{tabular*}{\linewidth}{@{}lllll@{}}
        \toprule
        methods & k-means & hierarchical & t-SNE &  DBSCAN\\
        \midrule
        DSC & 0.3633 & 0.3474 & 0.3716 & 0.3592 \\
        complexity & $O(n)$ & $O(n^2)$ & $O(n^2)$ & $O(n^2)$ \\
        \bottomrule
        \end{tabular*}
        \vspace{-1.5em}
    \end{table}
\end{small}

\textbf{Compare with Self-Supervised Learning Baselines}.
Then, we compared our MPS-AMS with state-of-the-art self-supervised methods on the BUSI, Hecktor, and BraTS2018 datasets with 5\% and 10\% labeled data. Firstly, we find that self-supervised methods generally outperformed fully supervised learning from scratch using partial annotations. This suggests that, in addition to limited labeled data, self-supervised methods also learn useful information from a large amount of unlabeled data. Secondly, when comparing MPS-AMS with SimCLR, BYOL, SwAV, MAE, and SimMIM, we observe that MPS-AMS significantly outperformed these methods on all datasets. Specifically, MPS-AMS achieves 5.18\%, 4.23\%, 7.51\%, 2.75\%, and 3.77\% higher DSC index scores than other self-supervised baselines under the 10\% annotation ratio. This improvement can be attributed to the use of masked patches selection and adaptive masking strategy in MPS-AMS, which reduces the uncertainty of masked patches and improves the upper bound of conditional mutual information, thereby learning more comprehensive and fruitful representation information.

Moreover, we have found that in some cases, the PPV of our method is not always the best. This may be attributed to marginal background patches that bear some resemblance to lesion patches. To address this issue, we plan to explore attention mechanisms~\cite{PAC,xu2019semi} to make model pay more attention to foreground patches.

\textbf{Analysis of Visualized Segmentation Results}.
Moreover, all the aforementioned findings are further supported by the visual results presented in Figure \ref{fig:visualization}. The results show that MPS-AMS outperforms all other self-supervised medical image segmentation methods, with segmentation results more similar to the ground truth. These visual observations provide further evidence that the effectiveness of our proposed masked patches selection and adaptive masking strategy.

\textbf{Analysis of Clustering Methods}.
We have tried different kinds of clustering methods. As shown in Table~\ref{table_clustering}, we can see that under the situation of Brats2018 with $10\%$ labeled data, the result of k-means is only $0.83\%$ different from the result of the best method t-SNE, but it can bring a significant reduction in computational complexity, thereby improving the efficiency of model operations, which indicates k-means can achieve good performance with lower complexity and faster efficiency. Considering actual hardware and time requirements, k-means is the best choice.

\begin{figure}[!t]
	\centering
        \includegraphics[width=0.3\textwidth]{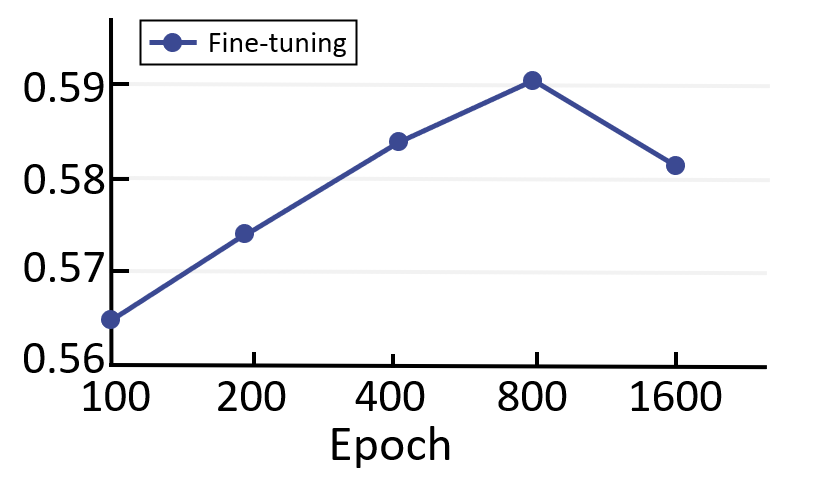}
        \vspace{-1.5em}
	\caption{Training schedules on BUSI with $10\%$ labeled data.}
        \vspace{-1.5em}
	\label{schedules}
\end{figure}

\textbf{Analysis of Training Schedules}.
We also test the performance of different epochs and the results are shown in Figure~\ref{schedules}. The vary trend before convergence is the same as MAE and it achieves the best when epoch reaches 800 as mentioned before. When the epoch continues to increase, performance begins to decline due to an excessive masking ratio.

\vspace{-1em}

\vspace{-0.3em}
\subsection{Ablation study}
\vspace{-0.5em}

To verify the effectiveness of our proposed MPS-AMS, we conducted further experiments to evaluate the segmentation performance of three intermediate models: \emph{base}, \emph{base+AMS}, and \emph{base+MPS}. The \emph{base} is a masked autoencoder based on U-Net, which aims to reconstruct masked patches of the input image and output an image of the same size as input image. Its masking ratio is 75\%. The \emph{base+AMS} model can be seen as the \emph{base} model with an adaptive masking strategy, while \emph{base+MPS} is the \emph{base} model based on the masked patches selection strategy. We conduct the above ablation studies on BUSI, Hecktor, and BraTS2018 datasets using 5\% ratios of annotations. The corresponding results are shown in Table~\ref{table:ablation_study}.

As shown in Table~\ref{table:ablation_study}, different strategies contribute differently to the model performance on segmenting tasks. Concretely, \emph{base+AMS} outperforms \emph{base}, indicating that fixed high masking ratio limits the upper bound of representation learning capacity and AMS can solve the problem effectively, \emph{base+MPS} outperforms \emph{base}, indicating the efficiency of masked patches selection strategy. Besides, MPS-AMS achieves the best performance when AMS and MPS are all utilized. For example, the DSC, PPV, and Sen increase $4.03\%$, $2.65\%$, and $6.85\%$ for the BUSI dataset. The results demonstrate that our proposed strategies are highly effective.

\vspace{-0.65em}
\section{Conclusion}
\label{sec:conclusion}
\vspace{-0.5em}

In this paper, we propose a self-supervised medical image segmentation method named MPS-AMS, which is based on masked patches selection and adaptive masking strategy. The proposed method can not only alleviate the limitations of current MIM methods in medical images, but also improve the upper bound of conditional mutual information, and reduce gradient noise, thus learning more representation information and achieving better performance. To evaluate the effectiveness of our method, we conduct extensive experiments on three public medical image datasets, and the results demonstrate that our method is effective in self-supervised medical image segmentation tasks.

Considering that there is abundant mutual information in multimodal medical image data and the imbalanced data problem, it is worth of conducting further investigations to apply MIM methods in multimodal~\cite{ZHANG-MIA2022} and imbalanced~\cite{wang2021rsg} medical image analysis tasks to extract more representation information and enhance the deep models' performances.

\vspace{-0.65em}
\section{Acknowledgments}
\vspace{-0.5em}

This work was supported by the National Natural Science Foundation of China under the grants 62276089, 61906063 and 62102265, by the Natural Science Foundation of Hebei Province, China, under the grant F2021202064, by the ``100 Talents Plan'' of Hebei Province, China, under the grant E2019050017, by the Open Research Fund from Guangdong Laboratory of Artificial Intelligence and Digital Economy (SZ) under the grant GML-KF-22-29, and by the Natural Science Foundation of Guangdong Province of China under the grant 2022A1515011474.

\vfill\pagebreak
\clearpage
    \bibliographystyle{IEEEbib}
    \small
    \bibliography{refs.bib}


\end{document}